\documentclass[runningheads]{llncs}

\usepackage{eccv}

\usepackage{eccvabbrv}

\usepackage{graphicx}
\usepackage{booktabs}

\usepackage[accsupp]{axessibility}  

\usepackage{hyperref}

\usepackage{orcidlink}

\begin{document}

\title{Delineate Anything v2: A Global Foundation Model for Field Delineation} 

\titlerunning{Delineate Anything v2}

\author{Mykola Lavreniuk\inst{1}\textsuperscript{*}\inst{,2}\orcidID{0000-0003-2183-8833} \and
Nataliia Kussul\inst{3}\orcidID{0000-0002-9704-9702} \and
Andrii Shelestov\inst{2,4}\orcidID{0000-0001-9256-4097} \and Yevhenii Salii\inst{2,4}\orcidID{0009-0006-0395-8099} \and Volodymyr Kuzin\inst{2,4}\orcidID{0009-0007-1077-0382} \and Charlotte Julia Li-Xing Wang\inst{1}\textsuperscript{*}\orcidID{0009-0007-0270-3470} \and Zoltan Szantoi\inst{1}\orcidID{0000-0003-2580-4382}}

\begingroup
\renewcommand\thefootnote{}
\footnotetext{\textsuperscript{*} International Research Fellow}
\endgroup

\authorrunning{M. Lavreniuk et al.}

\institute{European Space Agency, Frascati, Italy \and Space Research Institute NASU-SSAU, Kyiv, Ukraine \and University of Maryland, College Park, Maryland, USA \and National Technical University of Ukraine “Igor Sikorsky Kyiv Polytechnic Institute”, Kyiv, Ukraine}

\maketitle

\begin{abstract}

 Accurate agricultural field boundary delineation at large scale is a foundational task for food security, supply chain transparency, and carbon accounting. While vision foundation models like SAM show remarkable zero-shot capabilities, they frequently fail in geospatial domains due to topological complexity, cropland texturing patterns, and a lack of physical scale awareness. In this work, we introduce Delineate Anything v2, a globally scalable foundation model designed specifically for wide-area field boundary mapping. We construct FBIS-73M, a 73-million-instance multi-resolution dataset spanning 61 countries. To address the pervasive issue of multi-field administrative parcel merging, we introduce a resolution-specific data curation pipeline that leverages topological image-space adaptation to homogenize merged parcels and strengthen weak physical boundaries. Furthermore, we establish a novel, manually curated evaluation benchmark covering 100 countries to assess independent zero-shot generalization. Our results show that Delineate Anything v2 surpasses the current state-of-the-art, including the Delineate Anything framework, by 0.284 mAP@0.5 (+103.3\% relative gain), while maintaining execution speeds suitable for rapid national- and global-scale deployment, as demonstrated by nationwide mapping of Ukraine (603,000 km$^2$) in 5.4 hours on a consumer-grade workstation. Code, pre-trained weights, the FBIS-73M dataset, and ready-to-use national-scale vector boundary products are publicly available at \url{https://lavreniuk.github.io/Delineate-Anything/}.

\end{abstract}

\begin{figure}[t]
\centering
\includegraphics[width=\textwidth]{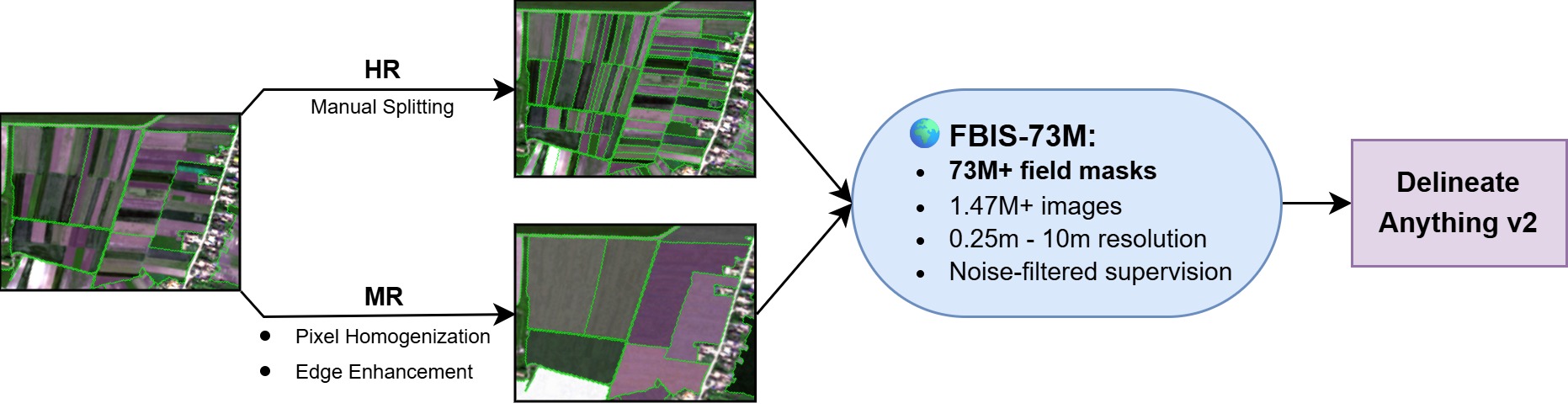}
\caption{Overview of the Delineate Anything v2 framework. To overcome structural label noise in global administrative registries, we introduce a resolution-aware data curation pipeline. Raw parcels undergo screening for anomalies. Problematic masks are then resolved via High-Resolution (HR) manual partitioning or Medium-Resolution (MR) topological image-space adaptation (pixel homogenization and edge enhancement). The purified FBIS-73M dataset is subsequently used to train the global foundation model.}
\label{fig:teaser}
\end{figure}

\section{Introduction}
\label{sec:intro}

Earth observation and geospatial artificial intelligence have entered an era of large-scale reasoning, driven by vision foundation models. Mapping the precise contours of agricultural fields serves as the primary spatial unit linking remote sensing to land-management decisions, crop-type identification, and agricultural statistics~\cite{Kussul2016, Kussul2017, Shelestov2017}. Accurate field boundaries are essential for assessing food security, monitoring deforestation compliance, and evaluating climate resilience in smallholder farming systems.

Historically, field boundary delineation relied either on classical supervised learning architectures trained on high-quality, local ground-truth (GT) datasets like the Land Parcel Identification Systems (LPIS), or on general-purpose vision foundation models like the Segment Anything Model (SAM)~\cite{Kirillov2023, ravi2024sam2segmentimages, carion2026sam3segmentconcepts}. While supervised models achieve strong local performance, they generalize poorly to regions lacking digital cadastral registries. Conversely, general foundation models struggle with satellite imagery due to low contrast between adjacent fields, sensor artifacts, and complex topologies, leading to frequent under-segmentation.

Despite recent advances in specialized architectures and datasets, benchmarks like FBIS-22M~\cite{Lavreniuk2025} remain concentrated in Europe, limiting geographic diversity across Africa, Asia, and Latin America. Furthermore, global scaling hits a major bottleneck: label noise in public administrative registries. Large-scale data aggregation suffers from the parcel-versus-field dilemma, where administrative boundaries group multiple physical fields into a single polygon. Automatically splitting these topologies is highly error-prone, limiting model accuracy. For global field delineation, data quality has become a more critical bottleneck than model architecture.

We introduce Delineate Anything v2, a globally representative geospatial foundation model driven by a data-centric paradigm. Instead of structural architectural changes, we systematically remediate label noise at the data level. Candidate merged parcels are identified through anomalous intra-mask spectral and textural variability. We then deploy a resolution-specific remediation strategy. Flagged high-resolution (HR) polygons are split manually for strict geometric accuracy. For medium-resolution (MR) samples, where manual correction is prohibitively expensive at scale, we introduce an image-space modification technique: anomalous regions are homogenized to erase false internal boundaries, while valid but visually weak physical edges are synthetically enhanced. This adapts image content to instance-level annotations without error-prone geometric label edits (Fig.~\ref{fig:teaser}).

The main contributions of this work are as follows:
\begin{itemize}
    \item We construct FBIS-73M, a 73-million-instance multi-resolution field boundary repository spanning 61 countries - the largest public dataset of its kind.
    \item We propose a data-centric noise remediation strategy that resolves the parcel-merging anomaly by combining manual partitioning for HR data with automated image-space pixel homogenization for MR data.
    \item We manually curate a high-fidelity validation benchmark spanning 100 countries to evaluate generalization across diverse agricultural regimes.
    \item We demonstrate that Delineate Anything v2 establishes a new state-of-the-art for global agricultural field delineation, surpassing the previous state-of-the-art Delineate Anything framework by 0.284 mAP@0.5 (+103.3\% relative gain) and enabling nationwide mapping of Ukraine (603,000 km$^2$) in 5.4 hours on a consumer-grade workstation.
\end{itemize}

\section{Related Work}
\label{sec:related_work}

\subsection{Operational Domain and Structural Requirements}

Agricultural field boundaries are the core spatial primitive for downstream operations, including yield estimation, crop classification, and environmental mapping. Governments also rely on parcel layers to validate subsidy claims~\cite{Scown2020}. Topological and geometric errors (under-segmentation, topology fragmentation) propagate into spatial statistics and compliance monitoring. Traditional edge detection, watershed segmentation, and object-based image analysis (OBIA)~\cite{Turker2013} scale poorly due to sensitivity to empirical parameter tuning across landscapes.

\subsection{Evolution of Deep Learning Architectural Paradigms}

Scalability limits motivated a shift toward deep learning, initially dominated by pixel-wise semantic segmentation backbones like U-Net and ResUNet-a~\cite{Diakogiannis2020, Waldner2020}. To resolve boundary ambiguity at field margins, research introduced specialized target formulations, such as multi-task learning architectures that simultaneously estimate boundary proximity, distance-to-boundary maps, and core interior masks~\cite{Waldner2020, Wang2021, Long2022}. Despite integrating advanced boundary-aware loss functions, semantic networks do not explicitly model instance-level awareness~\cite{Li2023}. This oversight results in topologically invalid outputs where adjacent parcels with identical spectral signatures are collapsed into single continuous objects~\cite{Pan2023}.

To prevent under-segmentation, recent methods adopt detection- and instance-segmentation architectures like Co-DETR~\cite{Zong2023} and real-time YOLO variants~\cite{Khanam2024, tian2025yolov12attentioncentricrealtimeobject, lei2025yolov13realtimeobjectdetection, jocher2026ultralyticsyolo26unifiedrealtime}, which enforce discrete object identity by construction, alongside backbone advances like ViT-Adapter~\cite{Chen2022} and EVP~\cite{Lavreniuk2025evp} that improve dense feature representations more broadly. While enforcing discrete object identity, these models remain bottlenecked by the lack of instance-level agricultural annotations.

\subsection{Geospatial Applications of Vision Foundation Models}

The inception of general-purpose foundation networks established new capabilities for zero-shot object segmentation. In remote sensing, SAM adaptations~\cite{Kirillov2023, ravi2024sam2segmentimages, carion2026sam3segmentconcepts} utilize prompt tuning~\cite{Osco2023, Ren2023, Chen2024, Luo2024}, multi-scale spatial attention~\cite{Feng2024, Long2024, Pan2025}, or weakly supervised boundaries~\cite{Sun2024, Wang2025}. However, general foundation models exhibit systematic failures in agricultural mapping: lacking geospatial awareness and scale constraints, they over-segment fields on minor crop variations or fail in low-contrast conditions with subtle boundaries~\cite{Huang2024}. 

\subsection{Data-Centric Scale and Label Noise Mitigations}

Dataset scaling limits constrain boundary extraction. AI4SmallFarms (439K instances, Vietnam/Cambodia)~\cite{Persello2023} or AI4Boundaries (2.5M parcels, Europe)~\cite{Dandrimont2023} datasets offer limited geographic coverage, while Fields of the World (1.6M parcels, 24 countries)~\cite{Kerner2024FTW} is restricted to 10m imagery.

Large-scale frameworks like Delineate Anything and FBIS-22M~\cite{Lavreniuk2025} expanded training data, but two limitations remain: FBIS-22M is concentrated in Europe, lacking representation in Africa, Asia, and Latin America; and administrative data aggregation introduces substantial label noise due to the mismatch between ownership parcels and physical field boundaries. Together, these limitations constrain the ability of current models to generalize reliably at a global scale.

Traditional approaches to label noise primarily focus on mitigating semantic corruption or wrong class assignments through robust optimization and label filtering techniques~\cite{Pelletier2017, Burgert2022, Hell2024}. While these methods are effective in many classification settings, they do not directly resolve the structural inconsistencies inherent in geospatial annotations~\cite{Liu2023, Liu2024}. Most existing approaches primarily address class-label corruption or annotation errors, but agricultural boundary datasets exhibit a different type of structural label noise, which arises from the mismatch between administrative parcels and physical cultivation units~\cite{Pan2024, Tang2025}.

Recent data-centric learning paradigms instead emphasize improving supervision quality at the source rather than compensating for annotation errors through increasingly sophisticated optimization strategies~\cite{Ng2021DataCentric, SambasivanShivaniKapaniaHannahHighfll2021, Zha2025}. Following this principle, Delineate Anything v2 explicitly targets label noise arising from the parcel-versus-field discrepancy. By identifying anomalous masks through intra-mask visual heterogeneity and applying resolution-specific remediation strategies, the framework improves consistency between image content and instance-level supervision, enabling reliable generalization across diverse agricultural systems and geographic regions.

\section{Methodology}
\label{sec:methodology}

The core philosophy of Delineate Anything v2 rests on data-centric curation. Rather than expanding model capacity through architectural modification, this framework scales and purifies the underlying training data to resolve systemic label noise characteristic of global administrative registries. 

\subsection{Dataset Statistics and Spatial Distribution}

\begin{figure}[!htbp]
\centering
    \includegraphics[width=\textwidth]{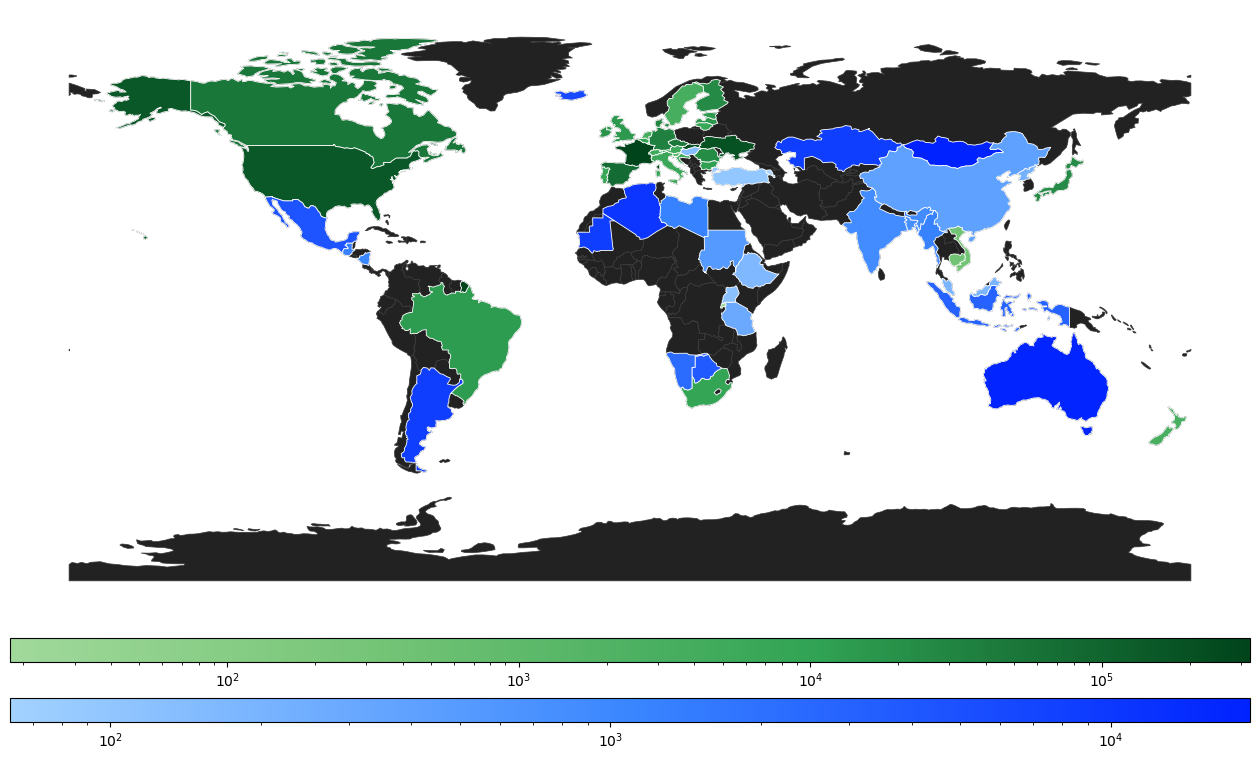}
    \vspace{0.3cm} \\
    \includegraphics[width=\textwidth]{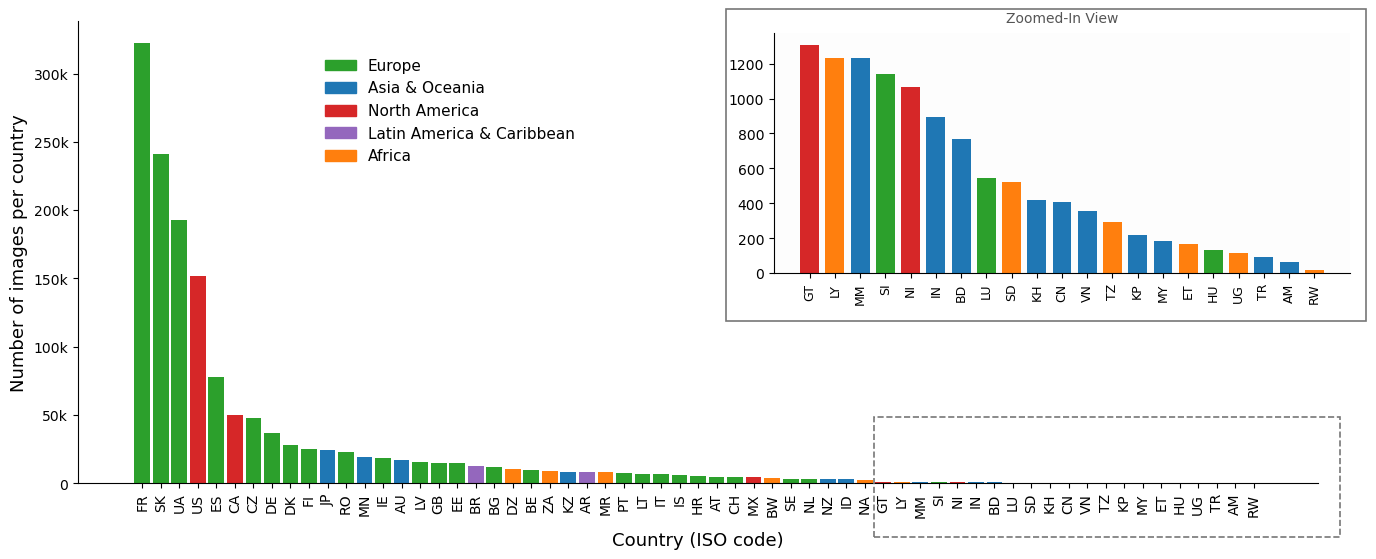}

\caption{Spatial distribution and country-level heterogeneity of the FBIS-73M training dataset. The global map (top) illustrates geographic sample density, where green shading corresponds to regions containing agricultural field instances (intensity proportional to patch count) and blue shading represents non-agricultural background reference countries. The histogram (bottom) quantifies the long-tailed sample distribution driven by public cadastral data availability across the 61 represented nations.}
\label{fig:train_dist_hist}
\end{figure}

To achieve a truly representative global model, we introduce two distinct data assets: the FBIS-73M training repository and a novel, independent 100-country evaluation benchmark. The FBIS-73M training dataset comprises 73~million discrete agricultural field instances distributed across 61 countries. The imagery spans multi-resolution satellite data ranging from high-resolution (HR, 0.25--3~m) to medium-resolution (MR, 3--10~m). Unlike prior datasets relying almost exclusively on European LPIS data, FBIS-73M spans multiple continents, ensuring robust global generalization (Fig.~\ref{fig:train_dist_hist}).

The training sample distribution is heterogeneous and long-tailed, reflecting open cadastral data availability rather than a deliberate sampling strategy (Fig.~\ref{fig:train_dist_hist}). Despite this imbalance, including low-resource regions gives FBIS-73M broader geographic diversity than prior Europe-centric datasets.

To assess zero-shot generalization independently, we created a manually annotated evaluation benchmark spanning 100 countries. Exactly four representative image patches per country were sampled to ensure equal geographic weighting and global coverage. The benchmark evaluates cross-country generalization rather than within-country variability; with equal patch counts per country, performance differences reflect landscape complexity rather than sampling density.

\begin{figure}[!htbp]
\centering

    \includegraphics[width=\textwidth]{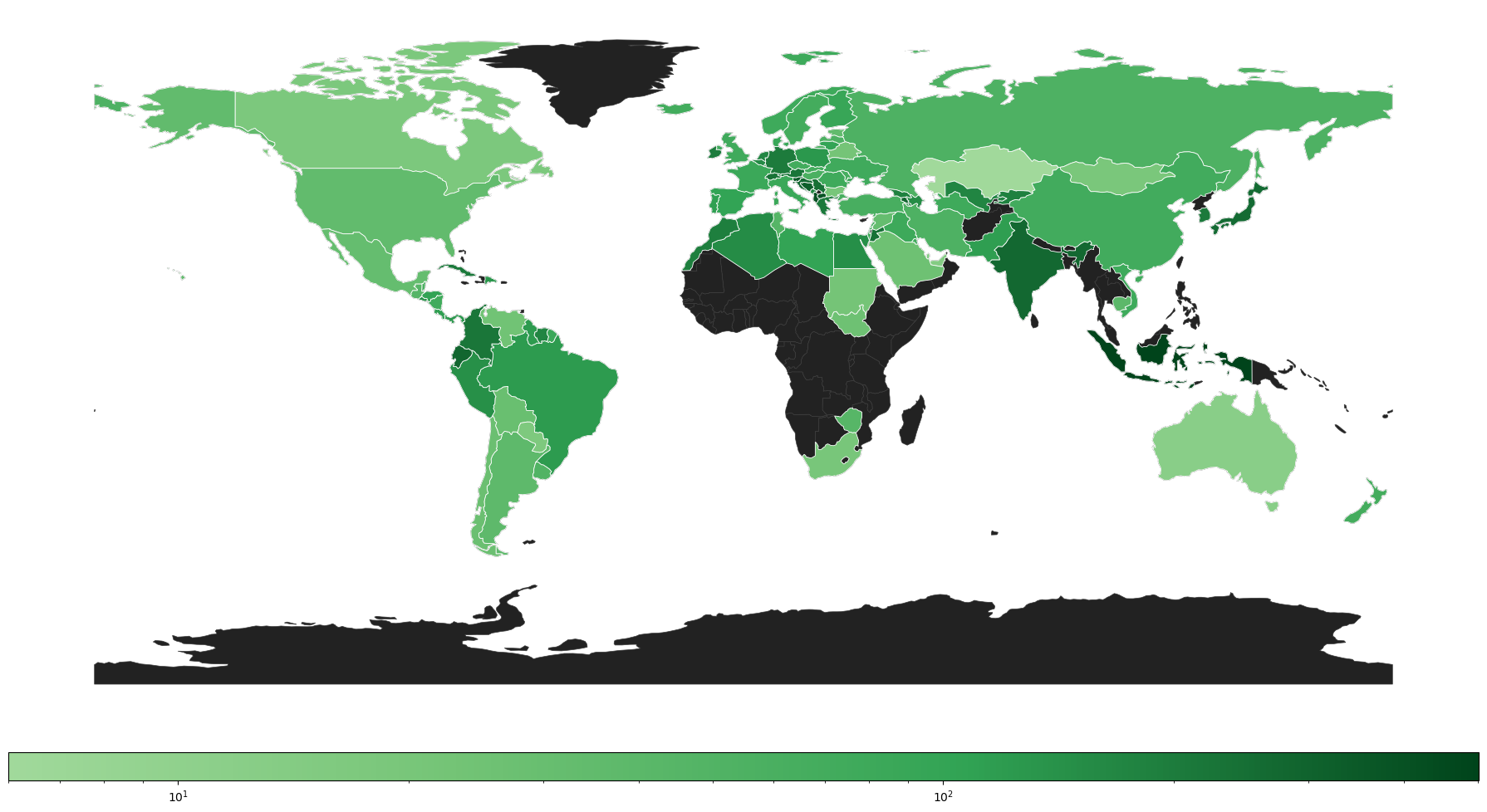}
    \vspace{0.3cm} \\
    \includegraphics[width=\textwidth]{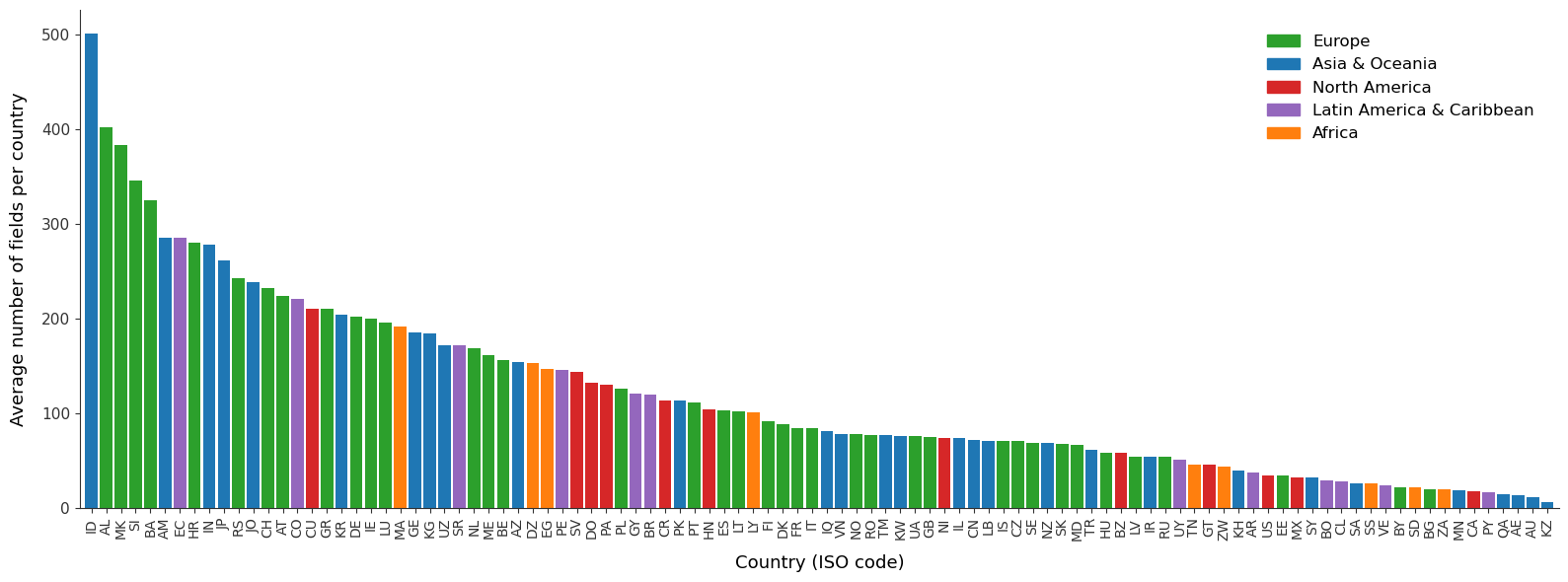}

\caption{Spatial and regional distribution of field density within the independent 100-country evaluation benchmark. The choropleth map (top) visualizes the geographic coverage of the test set (exactly 4 patches per country), with shading representing the average number of field instances per image. The distribution (bottom) quantifies the average number of fields per country, color-coded by continent.}
\label{fig:test_dist_hist}
\end{figure}

We therefore characterize each country by the average number of field polygons per image (Fig.~\ref{fig:test_dist_hist}). The resulting benchmark encompasses a wide spectrum of agricultural systems, ranging from large industrial fields in North America to highly fragmented smallholder landscapes in the Global South. In contrast to the geographically imbalanced but large-scale FBIS-73M training dataset, the evaluation benchmark prioritizes uniform country representation and structural diversity. Consequently, the benchmark assesses not only geographic transfer to previously unseen regions but also robustness to substantial variations in field size, fragmentation, density, and spatial organization.

However, aggregating global administrative data introduces two major challenges for model training. First, administrative parcels often merge multiple distinct physical fields into a single polygon. Second, true physical boundaries in satellite imagery frequently lack visual contrast, especially when adjacent fields contain similar crops or share identical phenological stages. To address both issues, we propose an automated data curation pipeline.

\subsection{Automated Screening for Label Noise}

Instead of relying on error-prone, geometric label modifications on a large scale, our pipeline begins with an automatic screening of the 73 million instances for intra-mask visual heterogeneity. Given an RGB image patch and its label raster, we suppress sensor noise via Gaussian blur ($13 \times 13$ kernel) and construct an eroded instance mask by retaining pixels where local minimum and maximum filtering yield identical semantic labels. This isolates the core of each administrative parcel, where we compute mean color values across blurred channels.

Candidate merged parcels are identified using a pixel-wise deviation map, assigning pixels to three deviation levels according to relative color difference from the parcel mean. To account for differing radiometric characteristics, baseline threshold sets are employed: $[10\%,20\%,50\%]$ for Sentinel-2 and $[25\%,35\%,50\%]$ for Planet imagery. These thresholds adaptively scale per instance based on its mean color amplitude.

After morphological erosion removes isolated noise, we compute the area of each deviation level. A parcel is flagged as a multi-field candidate if the cumulative area of the first, second, or third level exceeds 15\%, 10\%, or 5\% of the parcel area, respectively. The thresholds were selected conservatively to prioritize precision over recall and were kept fixed across all datasets.

\subsection{Image-Space Remediation: Pixel Homogenization}
\begin{figure}[!htbp]
\centering
\includegraphics[width=\textwidth]{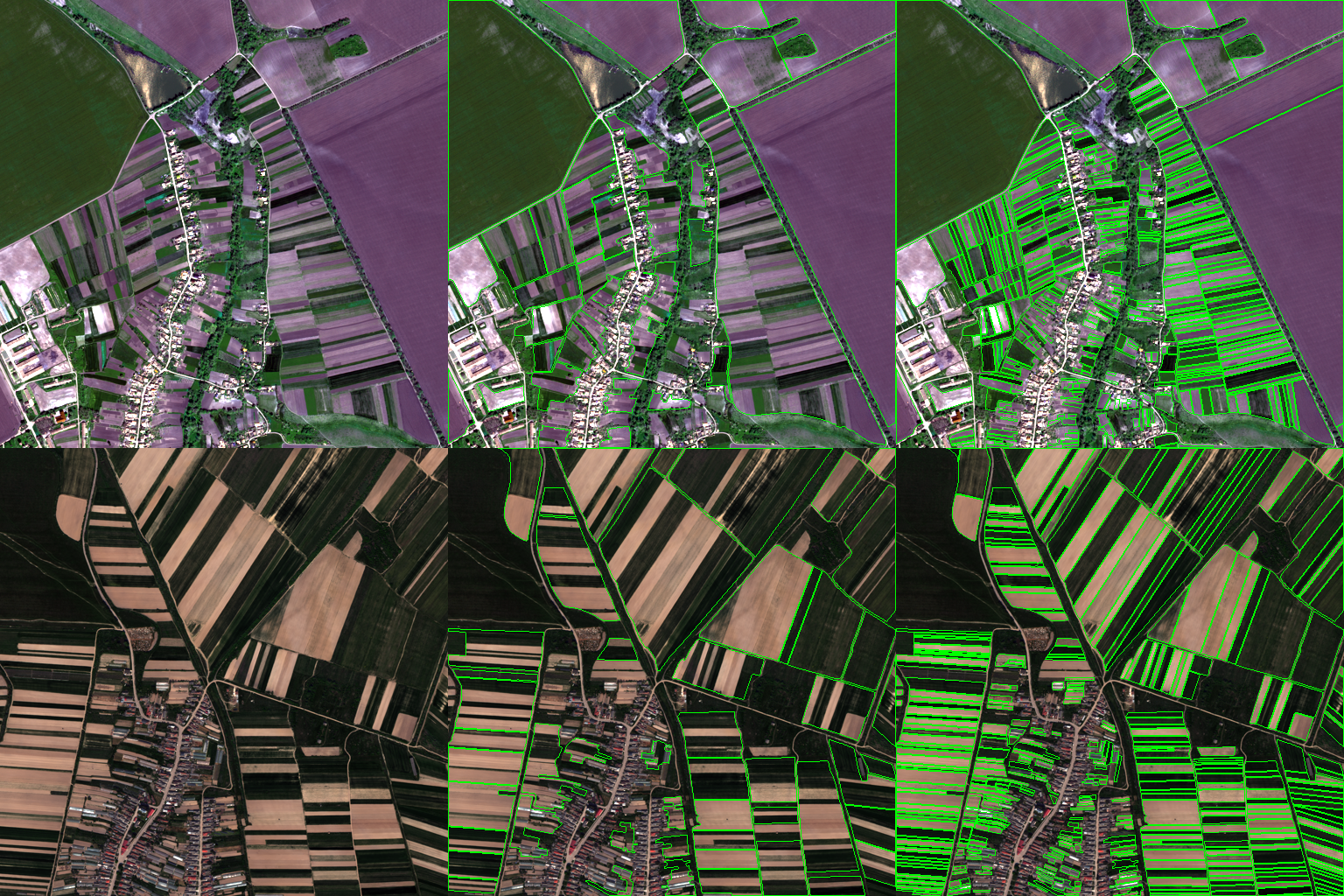}
\caption{Examples of high-resolution manual label refinement. The layout showcases two agricultural scenarios (rows) across three stages (columns, left to right): (1) raw satellite imagery, (2) original overlapping administrative mask with topological errors, and (3) finalized manually split vector boundaries matching the physical field structure.}
\label{fig:manual_curation}
\end{figure}

For high-resolution (HR) imagery ($\le 3$~m GSD), flagged parcels undergo targeted manual geometric decomposition into physical fields (Fig.~\ref{fig:manual_curation}). However, for medium-resolution (MR) imagery ($3-10$~m GSD), automated geometric splitting is prone to topological artifacts.

\begin{figure*}[!htbp]
\centering
\includegraphics[width=\textwidth]{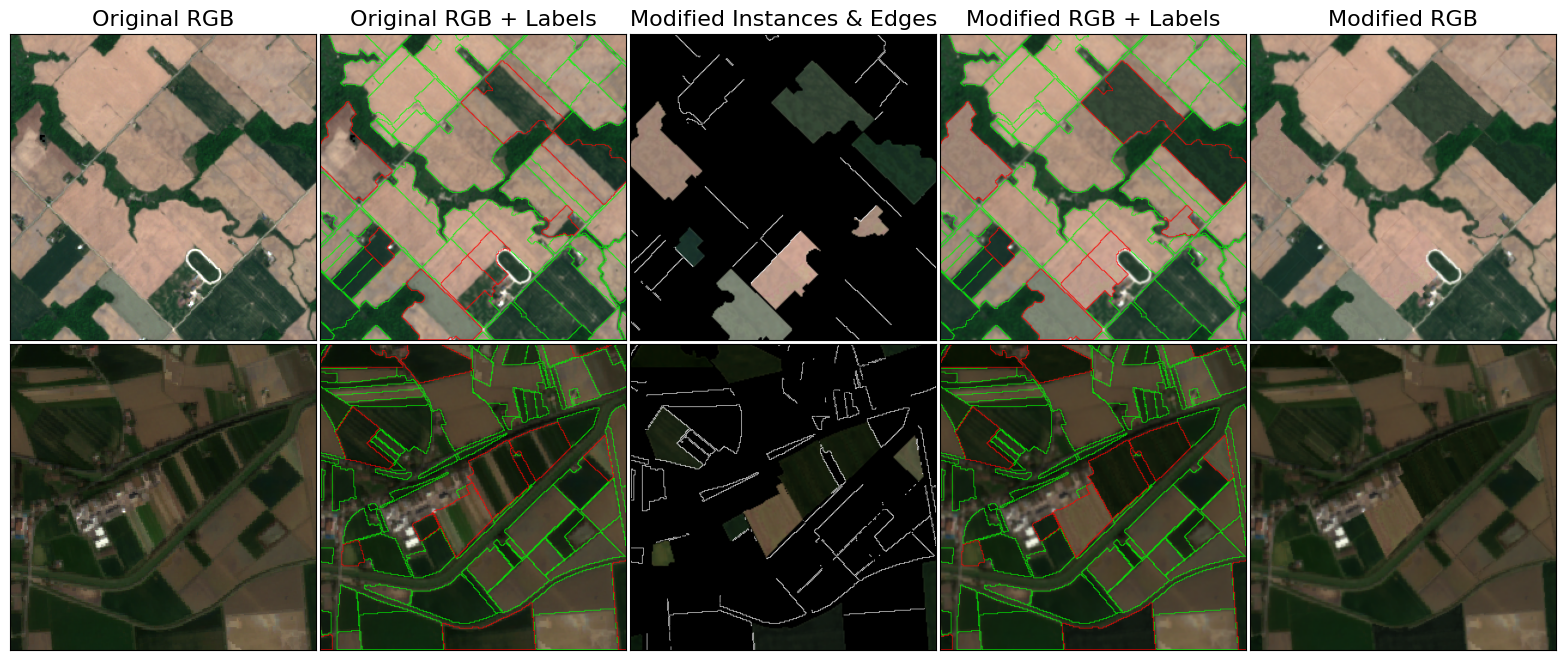}
\caption{Pipeline for automated medium-resolution topological image-space adaptation. The five panels show (left to right): (1) raw satellite patch, (2) original administrative contours with parcel-versus-field mismatches highlighted in red, (3) processed, homogenized fields and enhanced edges extracted via the anomaly mask, (4) modified raster with unaltered (green) and remediated (red) contours indicating where modifications occurred, and (5) the finalized, purified output patch.}
\label{fig:automated_remediation}
\end{figure*}

Instead, we introduce an image-space adaptation mechanism. Rather than dividing the label mask, we homogenize the underlying image pixels to match the single-instance supervision (Fig.~\ref{fig:automated_remediation}). We convert the image to the HSV color space, compute the mode color of the flagged instance, and calculate the deviation of the blurred HSV channels. To preserve textures while removing internal boundaries, we apply a cyclical bounding transformation to the deviations: 
\begin{equation}
\Delta' = \frac{B}{4} - \left| \left( (\Delta + \frac{B}{4}) \bmod B \right) - \frac{B}{2} \right|
\end{equation}
where $B$ denotes a sensor-specific bounding range that controls the maximum preserved color deviation. We empirically set $B=32$ for Sentinel-2 and $B=48$ for Planet imagery to account for their differing radiometric characteristics. 

For deviations satisfying $|\Delta| \leq B/4$, the transformation behaves approximately as an identity mapping, preserving natural pixel-level texture and local radiometric variability within the parcel. Larger deviations, which are more likely to correspond to sharp internal boundaries between distinct fields, are folded back into the bounded range. This reduces the discontinuity without collapsing the parcel into a uniform color. These bounded deviations are added to the mode color, converted back to RGB, and used to replace the original pixels within the flagged instance. This effectively erases false internal boundaries and ensures that the network is not penalized for predicting a single parcel when the raw image displays multiple fields.

\subsection{Topological Edge Enhancement via Khalimsky Grids}

While homogenization removes false internal boundaries, many true physical boundaries in MR satellite imagery lack strong visual contrast. To address this, we introduce a synthetic edge enhancement step based on digital topology.

To precisely localize inter-pixel boundaries, we map the image and label raster onto a Khalimsky grid of size $(2N+1) \times (2N+1)$, where $N$ is the dimension of the original image. In this topological space, odd-odd coordinates represent image pixels, odd-even/even-odd coordinates represent boundaries between two pixels, and even-even coordinates represent junctions~\cite{Khalimsky1990}.

We compute directional gradients by taking the difference between dilation and erosion along $1\times 4$ and $4\times 1$ kernels for horizontal and vertical edges, respectively, which guarantees non-negative magnitudes. Then, we take the maximum across orientation and color channel to obtain a single edge-strength map. These gradient magnitudes are mapped onto the edge elements of a Khalimsky grid constructed from the label raster, where edges between differently labeled pixels are marked as active. To isolate individual polygon sides, we identify junctions (even-even grid sites) where more than two active edges converge, indicating a corner shared by three or more polygons. We then zero out these junctions before applying connected-component labeling to split each polygon outline into discrete boundary segments. For each segment, the mean gradient strength and the fraction of low-gradient pixels are computed. Segments falling below an empirical strength threshold or exceeding a fraction-of-weak-pixels threshold are flagged as visually weak.

To strengthen weak boundaries, we construct a second Khalimsky grid where odd-odd sites hold original image pixels, and edge/junction sites contain the mean of neighboring pixels. For each such site, we compute a weighted brightness estimate and darken pixels brighter than a mid-range threshold (64 in 8-bit values) while lightening darker ones, applying a standard soft-light blending operation. The blending strength is governed by a parameter $\alpha \in [0,1]$ that interpolates between a neutral blend ($\alpha=0$) and a target contrast level ($\alpha=1$), allowing edge intensity to be tuned independently of the underlying image content. Finally, the enhanced grid is downsampled to original resolution by redistributing edge and junction deviations equally across neighboring pixels, injecting synthetic edge cues consistent with instance labels.

\section{Results}
\label{sec:results}

\subsection{Experimental Setup}
All experiments were conducted using a standardized training protocol to evaluate dataset scales and remediation variants fairly. Models were trained on $512 \times 512$ RGB image patches for 5 epochs using a cluster of 8$\times$ NVIDIA H100 GPUs with a total batch size of 512 (64 images per GPU). Optimization was performed via AdamW with a conservative learning rate of $1 \times 10^{-7}$ and a cosine decay schedule. The learning rate was selected empirically and used consistently across all experiments. Pre-trained COCO weights were used for initialization. To preserve natural field topologies, mosaic augmentation was strictly omitted, as it was found to introduce disruptive artifacts at patch boundaries; instead, standard flips, color jitter were applied. Performance is reported using standard COCO metrics~\cite{Lin2014}, specifically $\text{mAP@0.5}$ for coarse detection and $\text{mAP@[0.5:0.95]}$ for strict boundary precision, alongside precision and recall.

We adopt a YOLOv11-based instance segmentation backbone due to its optimal balance between boundary precision and high-throughput inference required for Earth-scale processing. While more recent iterations (YOLOv12~\cite{tian2025yolov12attentioncentricrealtimeobject}, YOLOv26~\cite{jocher2026ultralyticsyolo26unifiedrealtime}) were systematically evaluated, preliminary tests indicated that these configurations introduced optimization instability and offered no consistent benefits for complex geospatial topologies. Crucially, the proposed data-centric remediation pipeline remains entirely architecture-agnostic.

\subsection{Global and Regional Performance}

Table~\ref{tab:global_benchmark} reports quantitative results on the independent 100-country zero-shot benchmark. Delineate Anything v2 (DelAny v2) establishes a new state-of-the-art, doubling the original DelAny framework's mAP@0.5 (+0.284) and nearly tripling its strict mAP@0.5:0.95 metric (+0.175), alongside a substantial +0.294 gain in precision.

Table~\ref{tab:regional_results} details regional performance across major macro-regions, highlighting consistent global improvements. DelAny v2 effectively bridges the gap between western mechanized agriculture and highly fragmented smallholder systems. While absolute scores in Europe and North America exceed 0.61 mAP, historically under-represented regions demonstrate the most significant relative surges: performance doubles in Africa (+0.333) and nearly triples in Asia \& Oceania (+0.279). This confirms that geographic training diversity, when paired with strict supervision consistency, enables true planetary generalization.

\begin{table}[t]
\centering
\caption{Quantitative performance on the independent 100-country benchmark.}
\label{tab:global_benchmark}
\begin{tabular}{lcccc}
\toprule
Method & mAP@0.5 & mAP@0.5:0.95 & Precision & Recall \\
\midrule
DelAny & 0.275 & 0.103 & 0.345 & 0.454 \\
\textbf{DelAny v2} & \textbf{0.559} & \textbf{0.278} & \textbf{0.639} & \textbf{0.525} \\
\bottomrule
\end{tabular}
\end{table}

\begin{table}[t]
\centering
\caption{Regional performance breakdown (mAP@0.5) on the 100-country benchmark.}
\label{tab:regional_results}
\begin{tabular}{lcccccc}
\toprule
Method & Europe & Africa & Asia \& Oceania & Latin America & North America \\
\midrule
DelAny & 0.332 & 0.251 & 0.161 & 0.314 & 0.317 \\
\textbf{DelAny v2} & \textbf{0.612} & \textbf{0.584} & \textbf{0.44} & \textbf{0.563} & \textbf{0.618} \\
\bottomrule
\end{tabular}
\end{table}

\begin{table}[t]
\centering
\caption{Ablation study of dataset scaling and data-centric remediation strategies. }
\label{tab:ablation}
\begin{tabular}{lcc}
\toprule
Training Configuration & mAP@0.5 & Gains ($\Delta$) \\
\midrule
FBIS-22M (Baseline) & 0.275 & -- \\
FBIS-73M (Raw labels, No Curation) & 0.361 & +0.086 \\
FBIS-73M (+ HR Manual Split) & 0.409 & +0.048 \\
\textbf{Delineate Anything v2 (Full Curation Pipeline)} & \textbf{0.559} & \textbf{+0.150} \\
\bottomrule
\end{tabular}
\end{table}

\begin{figure*}[t]
\centering
\includegraphics[width=\textwidth]{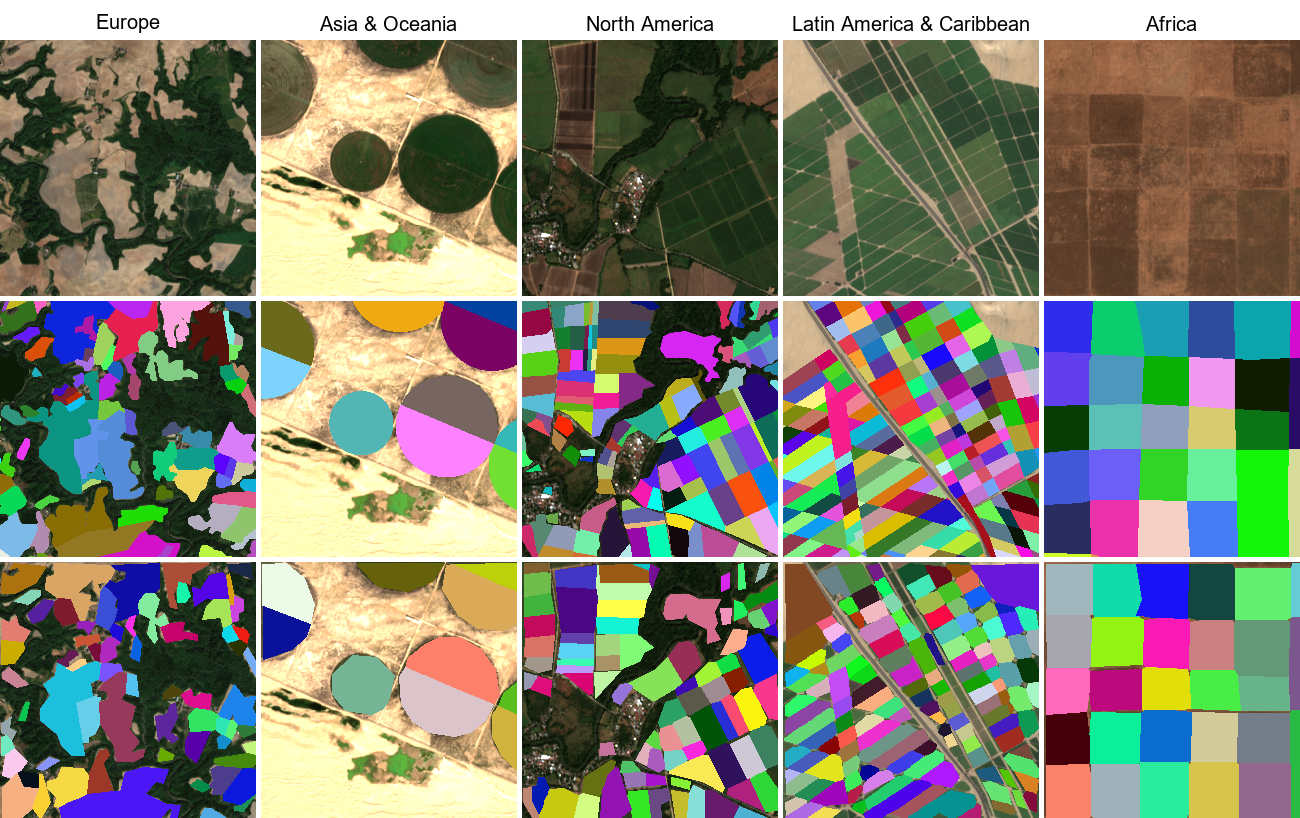}
\caption{Zero-shot qualitative results across major macro-regions. Rows (top to bottom): raw RGB imagery, manual ground truth, and Delineate Anything v2 predictions. Columns represent Europe (Norway), Asia \& Oceania (United Arab Emirates), North America (Costa Rica), Latin America \& Caribbean (Peru), and Africa (South Sudan).}
\label{fig:qualitative_predictions}
\end{figure*}

\subsection{Data-Centric Ablation Study}
To isolate the contribution of each curation component, we perform an incremental ablation study on the 100-country benchmark (Table~\ref{tab:ablation}). Scaling the raw training corpus from 22M to 73M instances yields a limited baseline expansion (+0.086), rapidly reaching a performance ceiling due to structural label noise.

Conversely, our data-centric interventions drive the majority of the overall accuracy gains. This further supports our central hypothesis that supervision quality is currently a larger bottleneck than dataset scale for global field boundary delineation. Incorporating High-Resolution (HR) manual geometric splitting provides a critical +0.048 mAP improvement. By resolving multi-field administrative parcel merging, this step effectively prevents topological collapse in highly fragmented smallholder fields. Finally, applying our automated Medium-Resolution (MR) image-space remediation (pixel homogenization and edge enhancement) contributes the largest individual performance surge (+0.150). Cumulatively, these data purification strategies contribute an additional +0.198 mAP over raw scaling, firmly establishing supervision quality, rather than raw dataset volume, as the primary bottleneck in foundational field delineation.

\subsection{Qualitative Analysis and Visual Benchmarking}
Figure~\ref{fig:qualitative_predictions} showcases zero-shot predictions across diverse continental ecozones, demonstrating the framework's capability to maintain sharp topological separation in ultra-fragmented landscapes and low-contrast boundaries. 

\begin{figure}[t]
\centering
\includegraphics[width=\textwidth]{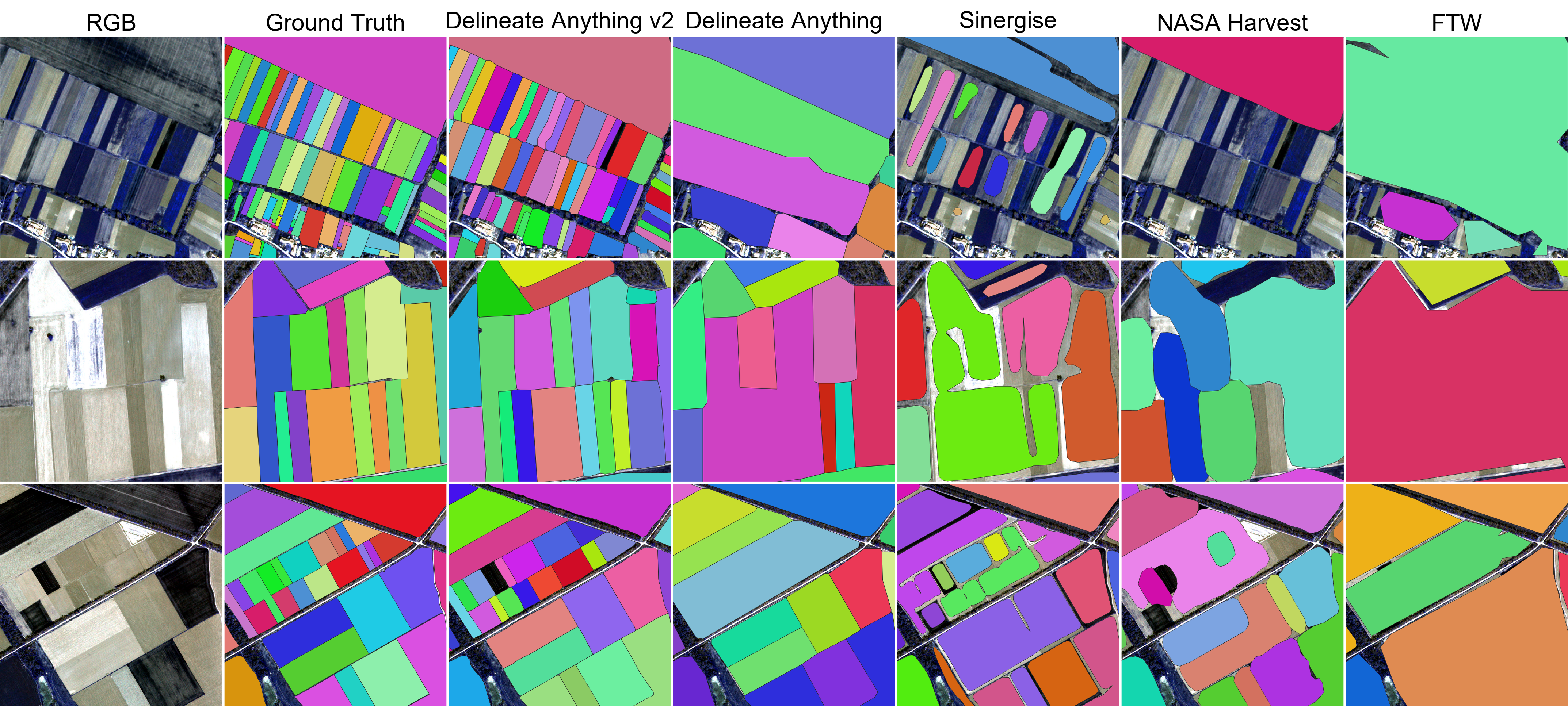}
\caption{Qualitative comparison against operational mapping products. The field boundaries extracted by Delineate Anything v2 and the manual ground truth are contrasted directly with publicly available static layers from Delineate Anything~\cite{Lavreniuk2025, lavreniuk2025delineateanythingflow}, NASA Harvest~\cite{Sadeh2025}, Sinergise Solutions~\cite{batic2024eu}, and Fields of the World (FTW)~\cite{kerner2025fields, muhawenayo2026prue}.}
\label{fig:product_comparison}
\end{figure}

Since direct metric benchmarking against baseline frameworks and open mapping pipelines is unfeasible without public weights~\cite{Sadeh2025, batic2024eu}, we utilize their static layers for localized visual comparison. As shown in Figure~\ref{fig:product_comparison}, existing wide-area layers (DelAny~\cite{Lavreniuk2025}, NASA Harvest~\cite{Sadeh2025}, Sinergise~\cite{batic2024eu}, FTW~\cite{kerner2025fields, muhawenayo2026prue}) frequently exhibit severe boundary collapse, completely omitting smallholder fields or merging small private plots into oversized administrative parcels. Conversely, DelAny v2 successfully recovers sharp, physically accurate field structures, demonstrating the clear advantage of training on topologically purified data.

\subsection{Operational Scalability}
Beyond benchmark evaluation, we assessed operational scalability by generating a complete 2024 field boundary map for Ukraine (603,000 km$^2$) using interpolated Sentinel-2 imagery. Delineate Anything v2 requires only 25 ms to process a single $512\times512$ image patch on an NVIDIA A100 GPU. End-to-end nationwide processing was completed in 5.4 hours on a consumer-grade workstation (AMD Ryzen 9 9900X CPU, NVIDIA RTX 5070 Ti 16 GB GPU, and 64 GB RAM), corresponding to approximately 112,000 km$^2$ processed per hour.

\section{Conclusion}
\label{sec:conclusion}

In this work, we introduce Delineate Anything v2, a foundation model for agricultural field boundary delineation, built on a data-centric supervision paradigm. While recent progress in geospatial vision has largely focused on architectural innovation and large-scale pretraining, our results show that label quality remains the fundamental bottleneck for field boundary extraction at global scale.

To address this, we construct FBIS-73M, a multi-resolution dataset of 73 million field instances across 61 countries, representing the largest publicly available field boundary repository to date. Beyond scaling data size, we propose a resolution-aware remediation pipeline that corrects the parcel-versus-field mismatch pervasive in public administrative registries.

For evaluation, we curate an independent benchmark covering 100 countries and diverse agricultural systems, enabling assessment of global generalization outside the training distribution. Experiments show that Delineate Anything v2 achieves a new state of the art in global field delineation, significantly outperforming prior methods and enabling nationwide mapping of Ukraine (603,000 km$^2$) in 5.4 hours on a consumer-grade workstation.

Overall, our findings suggest that progress in large-scale geospatial foundation models may depend less on increasing model complexity and more on improving the quality, consistency, and representativeness of supervision. We release FBIS-73M, the 100-country benchmark, model weights, code, and global vector products to support reproducible research in agricultural monitoring, food security, environmental compliance, and planetary-scale Earth observation.


\section*{Acknowledgements}
This work was partially supported by the European Space Agency (ESA), the European Commission through the joint World Bank/EU project ‘Supporting Transparent Land Governance in Ukraine’ [grant numbers ENI/2017/387-093 and ENI/2020/418-654], “Next-generation Copernicus services: an intelligent digital twin for forest cover analysis in Ukraine” (the Ministry of Education and Science of Ukraine), “DT4LC: Developing Scalable Digital Twin Models for Land Cover Change Detection Using Machine Learning” [grant number 2023.01/0040] (development and validation of training data), and NASA funded project “Detecting and Mapping War-Induced Damage to Agricultural Fields in Ukraine Using Multi-Modal Remote Sensing Data” (grant no. 80NSSC24K0354).

\bibliographystyle{splncs04}
\bibliography{main}
\end{document}